\documentclass[]{fairmeta}

\title{\vspace{-1pt}Fast LeWorldModel}

\author[]{Yuntian Gao}
\author[\dagger]{Xiangyu Xu}

\affiliation{Xi'an Jiaotong University}

\contribution[\dagger]{Corresponding author}

\newcommand{\method}{Fast-LeWM}
\newcommand{\fullmethod}{Fast LeWorldModel}
\newcommand{\obs}{\mathbf{o}}
\newcommand{\act}{\mathbf{a}}
\newcommand{\lat}{\mathbf{z}}
\newcommand{\pre}{\mathbf{p}}

\newcommand{\enc}{f_{\theta}}
\newcommand{\ape}{E_{\psi}}
\newcommand{\pred}{G_{\phi}}

\newcommand{\state}{\mathbf{s}}

\newcommand{\trainloss}{\mathcal{L}_{\mathrm{train}}}

\abstract{
\vspace{-5pt}
Joint-Embedding Predictive Architectures (JEPAs), including recent LeWorldModel (LeWM), have become a promising foundation for reconstruction-free visual world models.
For visual planning, however, LeWM evaluates candidate action sequences by repeatedly applying a local one-step latent transition model.
This autoregressive rollout makes planning computationally expensive and causes accumulated latent prediction errors as the horizon grows.
We propose \textbf{\fullmethod} (\method), a fast latent world model that replaces repeated local rollout with action-prefix prediction.
Given the current latent state and a candidate action sequence, \method{} encodes its prefixes and predicts the future latents reached after executing those prefixes in parallel.
By making action prefixes the basic prediction unit, \method{} directly models action effects accumulated over different horizons. 
Training these multi-horizon predictions jointly encourages the model to learn how states continuously evolve under different action prefixes, rather than only fitting one-step state transitions.
During planning, the predictor uses the prefix from the encoded action sequence to evaluate the corresponding future latent directly, without explicitly rolling through each intermediate imagined state.
Across multiple tasks, \method{} improves average success over LeWM while substantially reducing planning time. 
It also achieves lower open-loop latent prediction loss, with errors growing considerably more slowly as the rollout horizon increases.
\vspace{-8pt}
}

\metadata[Code]{\url{https://github.com/Yuntian-Gao/Fast-LeWorldModel}}
\metadata[Project Page]{\url{https://fast-lewm.github.io/}}
\usepackage{hyperref}
\usepackage{url}
\usepackage[T1]{fontenc}    %
\usepackage{url}            %
\usepackage{booktabs}       %
\usepackage{amsfonts}       %
\usepackage{nicefrac}       %
\usepackage{microtype}      %
\usepackage{epsfig, xspace, enumitem}
\usepackage{graphicx, amsmath, amssymb, caption, subcaption, multirow, overpic, textpos, pifont, adjustbox}

\usepackage{xcolor}
\usepackage[utf8x]{inputenc}
\makeatletter
\@namedef{ver@everyshi.sty}{}
\makeatother
\usepackage{pgf, tikz, pgfplots}
\usepackage{makecell}
\usepackage{pgf-pie}
\usepackage{float}
\usepackage{algorithm,algorithmic}
\usepackage{wrapfig}
\usepackage{colortbl}

\usepackage{newfloat}
\usepackage{listings}
\usepackage{booktabs}
\usepackage{siunitx}
\usepackage{multirow}
\usepackage{amsmath}
\usepackage{amssymb}
\usepackage{amsthm}

\theoremstyle{definition}

\theoremstyle{remark}

\setlist[enumerate]{itemsep=-0.5mm,partopsep=0pt}

\pgfplotsset{compat=1.18}
\begin{document}

\maketitle

\begin{figure}[!htb]
  \centering
  \includegraphics[width=0.85\textwidth]{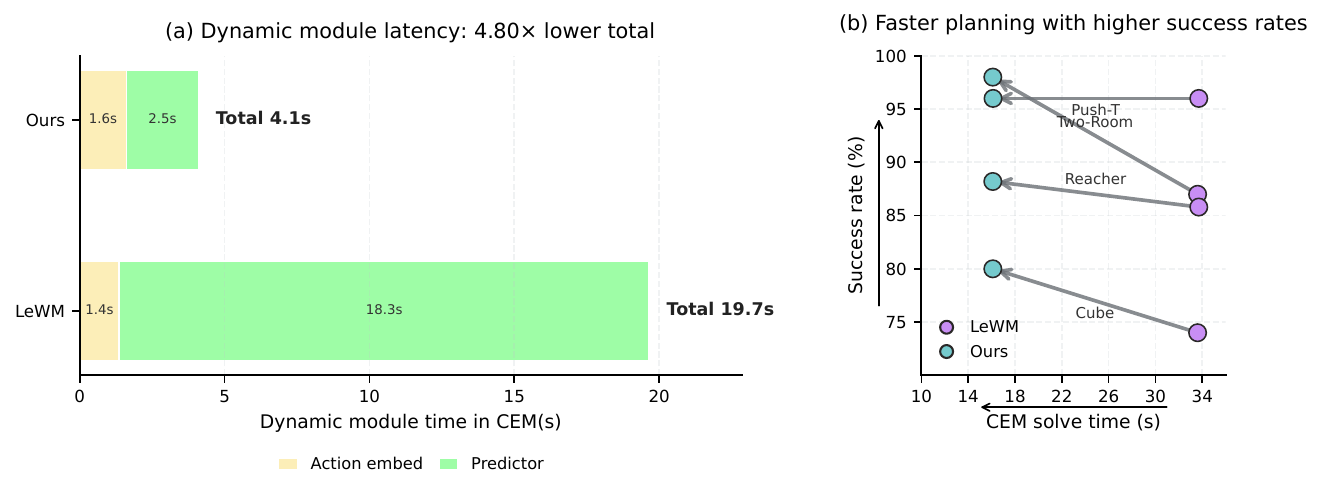}
\vspace{-2mm}
  \caption{
    Planning efficiency and task success rates.
    (a) Runtime breakdown of the two dominant rollout modules. 
    By replacing autoregressive latent rollout with parallel prefix prediction, our method reduces the dynamics-module time in CEM, including action encoding and latent prediction, from 19.7 to 4.1s.
    (b) Full CEM planning results on four tasks.
    Each arrow connects LeWM~\citep{maes2026leworldmodel} and our method under the same planning protocol, showing that our method consistently reduces CEM solve time from about 33.7s to 16.1s while improving task success rate.
}   
\label{fig:planning_time_breakdown}
\end{figure}
\vspace{-2mm}

\section{Introduction}
World models enable agents to plan by predicting the consequences of their actions before acting~\citep{ha2018recurrent,hafner2019learning,hafner2020dream, schrittwieser2020mastering, wu2023daydreamer}.
For visual planning, reconstruction-free JEPA-style world models learn to predict future embeddings rather than pixels, and recent systems such as LeWorldModel (LeWM) show that latent prediction can support reward-free goal-conditioned planning~\citep{lecun2022path,assran2023ijepa,bardes2024vjepa, assran2025v, sobalstress,zhoudino,maes2026leworldmodel}.

Despite this progress, LeWM-style planning remains costly and susceptible to accumulated prediction error.
Its dynamics model is inherently local: it predicts only the next latent state $\hat{\mathbf{z}}_{t+1}$ from the current latent $\mathbf{z}_t$ and action $\mathbf{a}_t$.
To evaluate a candidate action sequence, the planner must therefore apply this one-step model autoregressively, repeatedly feeding predicted latents back into the dynamics model $G_{\mathrm{LeWM}}$, and the terminal prediction $\hat{\mathbf{z}}_{t+H}$ is obtained by composing local transitions:
\[
\hat{\lat}_{t+H}
=
G_{\mathrm{LeWM}}\!\left(
G_{\mathrm{LeWM}}\!\left(
\cdots
G_{\mathrm{LeWM}}(\lat_t,\act_t),
\ldots
\right),
\act_{t+H-1}
\right).
\]

where each prediction depends on the intermediate imagined latents produced along the rollout.

This local transition interface creates two practical limitations.
First, candidate evaluation becomes slow because autoregressive rollout must generate the entire imagined latent trajectory step by step, repeating action encoding and latent prediction. 
Second, errors introduced at early rollout steps can propagate into later predictions, making rollout increasingly unreliable as the horizon grows.

We propose \textbf{\fullmethod} (\method), a fast latent world model built around action-prefix prediction. Instead of advancing the latent state one step at a time, \method{} treats an action prefix as the basic unit for querying future dynamics.
Given the current visual latent \(z_t\) and an action prefix
\(
\act_{t:t+k-1}
=
(\act_t,\ldots,\act_{t+k-1}),
\)
\method{} predicts the future latent reached after executing that prefix:
\[
\hat{\lat}_{t+k}
=
G_{\mathrm{Fast\text{-}LeWM}}
\!\left(
\lat_t,\act_{t:t+k-1}
\right),
\qquad
k=1,\ldots,H.
\]
Rather than estimating a long trajectory through repeated adjacent transitions, \method{} directly predicts the latent outcomes of different action prefixes from the observed anchor latent $\mathbf{z}_t$.
Different prefixes contain different degrees of accumulated action effects and correspond to different future latents, so prefix prediction gives the dynamics model a direct interface for multi-horizon state evolution.
This formulation also makes each prediction horizon directly supervisable from the same observed anchor latent, enabling dense multi-horizon learning without recursively feeding predicted latents back into the model.

To implement this interface efficiently, \method{} uses an action-prefix encoder and a parallel latent predictor.
The action-prefix encoder processes a candidate action sequence with a causal mask and converts it into multiple prefix tokens, where the token at position \(k\) summarizes only \(\act_t,\ldots,\act_{t+k-1}\) together with the current latent context.
A parallel predictor then maps the current latent and all prefix tokens to their corresponding future latents in one forward pass.
Dense prefix-level supervision ties each prefix token to the future latent reached after executing the corresponding partial action sequence.
Through this training objective, the model learns to predict in parallel how the state evolves under different action prefixes, rather than only modeling local one-step state evolution.

This design naturally benefits planning.
At planning time, the predictor takes all action prefixes as input and predicts all corresponding future states in parallel. 
These state predictions do not depend on one another sequentially, so errors are not recursively propagated through the $H$ intermediate predicted latents.
This reduces recursive error propagation within the planning horizon and accelerates state-evolution prediction through parallel prefix prediction.
As shown in Fig.~\ref{fig:planning_time_breakdown}, \method{} substantially reduces the dynamics-module cost inside the Cross-Entropy Method (CEM)~\citep{rubinstein2004crossentropy} while improving or preserving task success.
The proposed method also lowers open-loop prediction error and slows its growth over longer horizons.

Our contributions are:
\begin{itemize}
    \item We identify the local one-step transition interface in LeWM as a key bottleneck, leading to slow autoregressive rollouts and the accumulation of latent prediction errors over long horizons.
    \item We propose \method{}, a fast latent world model that reformulates latent dynamics modeling from single-step transitions to action-prefix prediction, which enables dense prefix-level training and turns accumulated action effects into direct dynamic training targets.
    \item We evaluate \method{} on all planning tasks from LeWM under the same protocol. \method{} improves the average success rate from 85.8\% to 90.5\%, accelerates the dynamics module by 4.8$\times$ (19.7s to 4.1s), reduces the full CEM solve time by 52.2\% (33.7s to 16.1s), and substantially lowers both open-loop prediction error and its growth over the planning horizon.
\end{itemize}

\section{Related Work}

\noindent\textbf{Latent World Models for Planning.}
World models learn action-conditioned dynamics for planning~\citep{ha2018recurrent,hafner2019learning,hafner2020dream,hou2026world}.
PlaNet and Dreamer learn latent dynamics from images for planning and behavior learning~\citep{hafner2019learning,hafner2020dream,hafner2023mastering}, while TD-MPC and TD-MPC2 use decoder-free models for continuous control~\citep{hansen2022tdmpc,hansen2024tdmpc2}.
In offline reward-free settings, CEM optimizes action sequences by matching predicted future latents to a goal~\citep{rubinstein2004crossentropy,sobalstress,zhoudino,maes2026leworldmodel}.
Repeated candidate evaluation makes rollout speed and long-horizon accuracy critical.

\noindent\textbf{Reconstruction-Free Visual Dynamics.}
JEPAs predict future or masked embeddings instead of reconstructing pixels~\citep{lecun2022path,assran2023ijepa,bardes2024vjepa}.
For visual planning, PLDM learns from reward-free trajectories with collapse-prevention regularization~\citep{bardes2022vicreg,sobal2022joint,sobalstress}, while DINO-WM predicts frozen DINOv2 patch features~\citep{oquab2024dinov2,zhoudino,yin2026ddp}.
LeWM trains an end-to-end JEPA world model using next-embedding prediction and SIGReg-style Gaussian regularization~\citep{balestriero2025lejepa,maes2026leworldmodel}; recent work further develops this regularization~\citep{zhao2026sub,thil2026subspace,wu2026visreg}.
Our contribution instead concerns the dynamics-query interface:  \method{} uses action prefixes as multi-horizon queries, conditioning each future latent on the observed anchor latent and the corresponding action prefix rather than solely on the immediately preceding imagined latent, which mitigates the compounding dependence of sequential rollout.

\begin{figure}[t]
  \centering
  \includegraphics[width=0.92\textwidth]{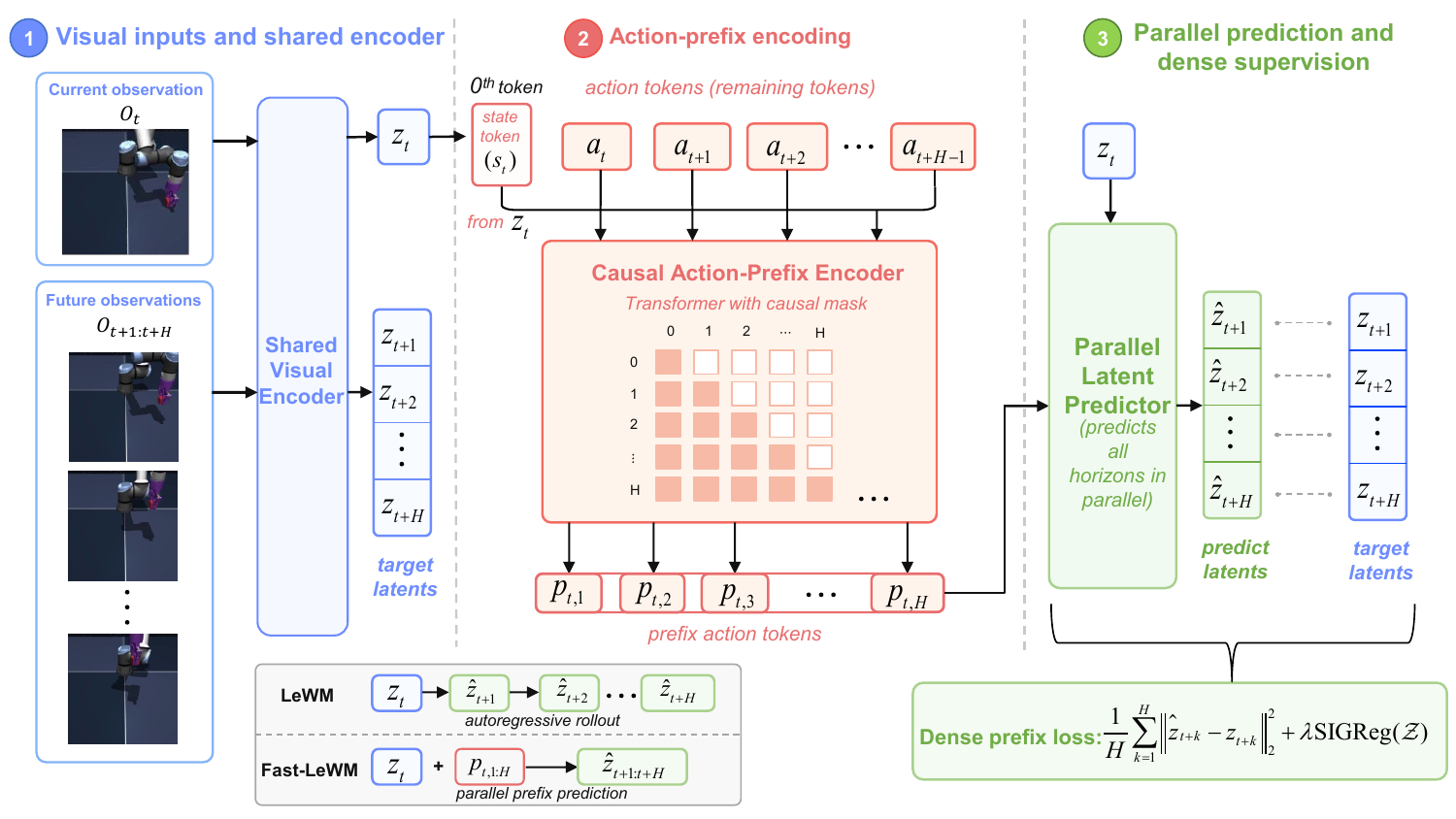}
  \vspace{-2mm}
\caption{\textbf{\method{} training pipeline.}
  The current observation $o_t$ is encoded into a latent state $z_t$.
  An Action-Prefix Encoder takes a state-action token sequence formed by prepending a state token mapped from $z_t$ as the 0-th token to the action tokens $(a_t,\ldots,a_{t+H-1})$.
  Under causal masking, the action-position outputs form prefix tokens $(p_{t,1},\ldots,p_{t,H})$, where $p_{t,k}$ corresponds to the prefix $(a_t,\ldots,a_{t+k-1})$, while the 0-th output token is discarded.
  Using the prefix tokens together with the anchor latent, the Parallel Predictor directly generates all future latents $(\hat{z}_{t+1},\ldots,\hat{z}_{t+H})$ in one forward pass, supervised by encoded future observations through a dense multi-horizon loss.
  Unlike isolated one-step transition learning, dense action-prefix prediction supervises multiple horizons directly, forcing the model to learn the multi-horizon state evolution induced by an action sequence while reducing error propagation into future steps.}
  \label{fig:action_prefix_pipeline}
  \vspace{-3mm}
\end{figure}

\section{Method}
\subsection{Reward-Free Latent World Models}

We consider an offline, reward-free dataset of observation-action trajectories
\begin{equation*}
    \mathcal{D}
    =
    \{\tau^{(n)}\}_{n=1}^{N},
    \qquad
    \tau
    =
    \{(o_t,\act_t)\}_{t=1}^{T}, 
\end{equation*}
where $o_t$ is a pixel observation and $\act_t \in \mathbb{R}^{d_a}$ is a continuous action.
The goal is to learn a latent world model that supports goal-conditioned planning without reward labels.
A visual encoder maps observations into latent embeddings $\lat_t = f_\theta(o_t) \in \mathbb{R}^{d}$,
and a goal observation $o_g$ is similarly encoded as $\lat_g=f_\theta(o_g)$.

At test time, the planner evaluates candidate action sequences by predicting the latent states that they would reach from the current state.
Each candidate is then assigned a planning cost according to the discrepancy between its predicted future latent and the goal latent $\lat_g$.
The dynamics model therefore serves as the interface that maps a candidate action sequence to the future latent representation used for planning.

\subsection{Autoregressive Rollout in LeWorldModel}

A common latent world model design, adopted by LeWM, learns a local one-step transition predictor $\hat{\lat}_{t+1}
    =
    F_\phi(\lat_t,\act_t),$
which is trained with a next-latent prediction loss:
\begin{equation*}
    \mathcal{L}_{\mathrm{pred}}^{\mathrm{1step}}
    =
    \left\|
    \hat{\lat}_{t+1}
    -
    \lat_{t+1}
    \right\|_2^2. 
\end{equation*}
Following reconstruction-free JEPA-style world models, an anti-collapse regularizer, such as SIGReg~\citep{maes2026leworldmodel}, is applied to a batch of latent embeddings $Z$:
\begin{equation*}
    \mathcal{L}_{\mathrm{1step}}
    =
    \mathcal{L}_{\mathrm{pred}}^{\mathrm{1step}}
    +
    \lambda\,\mathrm{SIGReg}(Z). 
\end{equation*}

Given a candidate sequence $\act_{t:t+H-1}$, the one-step model estimates the future trajectory by autoregressive rollout,
\begin{equation*}
    \hat{\lat}_{t+k}
    =
    F_\phi(\hat{\lat}_{t+k-1}, \act_{t+k-1}),
    \qquad
    k=1,\ldots,H, 
\end{equation*}
with initialization \(\hat{\lat}_{t}=\lat_t\).
Thus, obtaining the terminal prediction \(\hat{\lat}_{t+H}\) requires sequentially generating all intermediate latent states
\(\hat{\lat}_{t+1},\ldots,\hat{\lat}_{t+H-1}\).
Next, for goal-conditioned planning,
the candidate sequence is assigned a terminal latent-space cost:
\begin{equation}\label{eq:CAR}
    C_{\mathrm{AR}}(\act_{t:t+H-1})
    =
    \left\|
    \hat{\lat}_{t+H}
    -
    \lat_g
    \right\|_2^2. 
\end{equation}

The main drawback of this interface is the long autoregressive chain required during candidate evaluation.
Although the planning cost in Eq.~\ref{eq:CAR} depends only on the terminal latent,
reaching those states requires generating every preceding imagined latent and repeatedly invoking the action encoder and latent transition modules.
This repeated rollout becomes expensive when CEM evaluates a large population of candidate sequences over multiple optimization iterations.
Moreover, autoregressive rollout can also amplify prediction errors over long horizons.
Each intermediate prediction is reused as the input to the next transition, causing approximation errors introduced at early steps to accumulate through all subsequent predictions.

\subsection{Fast LeWorldModel}

We propose \method{} to address these issues. %
Instead of producing $\hat{\lat}_{t+1},\ldots,\hat{\lat}_{t+H}$ through an autoregressive chain, our key idea is to predict each future state directly from the observed anchor latent and the action prefix that leads to that state.
Under this view, different future states are not generated by feeding one predicted state into the next; they can be queried independently once their corresponding action prefixes are encoded.
This motivates the design of an action-prefix encoder together with a parallel latent predictor.
Fig.~\ref{fig:action_prefix_pipeline} gives an overview of the resulting training pipeline.

For a candidate action sequence $\act_{t:t+H-1}$, each horizon $k$ corresponds to the prefix $\act_{t:t+k-1}$.
The target associated with this prefix is the latent state reached after executing these $k$ actions, namely $\lat_{t+k}$.
\method{} maps each prefix to a learned prefix token
\begin{equation*}
    \pre_{t,k}
    =
    \ape^{(k)}(\act_t,\ldots,\act_{t+k-1}), 
\end{equation*}
where $\pre_{t,k}$ summarizes the accumulated effect of the first $k$ actions.
The latent predictor then uses the anchor latent $\lat_t$ and each prefix token $\pre_{t,k}$ to predict the corresponding future latent $\hat{\lat}_{t+k}$. %

This design has two direct benefits.
During training, dense supervision over all prefix tokens forces the model to learn how action effects accumulate over different horizons, rather than only fitting local one-step state evolution.
During rollout, all action prefixes can be processed together and their corresponding future states can be generated in parallel.
Because these predictions are anchored at the observed latent $\lat_t$ and do not depend sequentially on one another, the model reduces recurrent error accumulation while accelerating rollout through parallel state-evolution prediction.

\subsection{Action-Prefix Encoder}\label{sec:action-prefix encoder}

The role of the action-prefix encoder is to output horizon-specific prefix tokens $\pre_{t,1:H}$, where each token \(\pre_{t,k}\) encodes only the first \(k\) actions of the sequence and cannot access any subsequent actions.
Given an action sequence, the encoder outputs a dense sequence of prefix tokens
\begin{equation*}
    \pre_{t,1:H}
    =
    \ape(\act_{t:t+H-1})
    =
    (\pre_{t,1},\pre_{t,2},\ldots,\pre_{t,H}). 
\end{equation*}
We instantiate $\ape$ as a causal Transformer over action tokens.
With the causal mask, the representation at horizon $k$ can attend only to actions $\act_t,\ldots,\act_{t+k-1}$.
Thus, $\pre_{t,k}$ is aligned with the target $\lat_{t+k}$ and does not depend on future actions $\act_{t+k},\ldots,\act_{t+H-1}$.

In practice, an action prefix does not determine its outcome by itself.
The same open-loop actions may move the agent toward different regions, produce different contacts, or have different object effects depending on the current configuration of the scene. To provide this context with minimal overhead, we map the current latent through a lightweight MLP into a state token and prepend it to the action-token sequence before the causal Transformer as shown in Fig.~\ref{fig:action_prefix_pipeline}. %

\subsection{Parallel Latent Predictor}

Given the current latent and prefix tokens, the predictor estimates future latents for all prefix horizons in parallel:
\begin{equation*}
    \hat{\lat}_{t+1:t+H}
    =
    \pred(\lat_t,\pre_{t,1:H}). 
\end{equation*}
Specifically, the prediction at horizon \(k\) is given by
\begin{equation}
    \hat{\lat}_{t+k}
    =
    \pred(\lat_t,\pre_{t,k}),
    \qquad
    k=1,\ldots,H. \label{eq:predictor}
\end{equation}
Here, each prefix token \(\pre_{t,k}\) conditions the predictor on the cumulative effect of executing the first \(k\) actions, while the current latent \(\lat_t\) serves as a shared anchor for all prediction horizons.
The model therefore learns to map action prefixes of different lengths directly to their corresponding future latent states. %
This design is computationally favorable for planning: all queried horizons share one action-prefix encoding pass and one parallel latent-prediction pass, without repeatedly iterating one-step transitions. %

\subsection{Dense Prefix Prediction Objective}

For a training segment 
\begin{equation*}
    (o_t,\act_t,o_{t+1},\act_{t+1},\ldots,\act_{t+H-1},o_{t+H}), 
\end{equation*}
we encode the current and future observations as
\begin{equation*}
    \lat_{t+i}
    =
    f_\theta(o_{t+i}),
    \qquad
    i=0,\ldots,H. 
\end{equation*}
The action-prefix encoder and predictor produce prefix-level predictions $\hat{\lat}_{t+1:t+H}$.
Every prediction receives its own latent target ${\lat}_{t+1:t+H}$, leading to our loss function:
\begin{equation*}
    \mathcal{L}_{\mathrm{prefix}}
    =
    \frac{1}{H}
    \sum_{k=1}^{H}
    \left\|
    \hat{\lat}_{t+k}
    -
    \lat_{t+k}
    \right\|_2^2. 
\end{equation*}
This dense objective supervises not only the terminal outcome but also the intermediate states induced by partial action prefixes, forcing the model to learn how the latent state evolves as more actions are appended to the sequence.

We retain the SIGReg regularizer used by reconstruction-free latent world models to prevent latent collapse:
\begin{equation*}
    \mathcal{L}_{\mathrm{AP}}
    =
    \mathcal{L}_{\mathrm{prefix}}
    +
    \lambda\,\mathrm{SIGReg}(Z). 
\end{equation*}

\subsection{Planning with Action Prefixes and Self-Consistency}

During planning, we use the CEM-based goal-conditioned planning framework similar to LeWM.
Let $m$ index a candidate action sequence sampled by CEM, LeWM reaches $\hat{\lat}_{t+H}^{(m)}$ by repeatedly applying a one-step dynamics model along imagined latents.
In contrast, \method{} treats the action prefix as the rollout unit, so the corresponding prefix token $\pre_{t,H}^{(m)}$ provides a direct path from the current latent to that future latent.

The candidate is evaluated according to its terminal latent distance from the goal:
\begin{equation*}
    C_{\mathrm{goal}}^{(m)}
    =
    \left\|
    \hat{\lat}_{t+H}^{(m)}
    -
    \lat_g
    \right\|_2^2. 
\end{equation*}
CEM uses the candidates with the lowest costs to update its sampling distribution,
and the selected action sequence is executed until the next decision point.

This prefix-level formulation also enables an optional self-consistency signal during planning.
In addition to predicting the terminal latent directly from the length-$H$ prefix, the model can take one intermediate prefix step and then predict the remaining horizon from the intermediate latent, yielding another terminal estimate $\tilde{\lat}_{t+H}^{(m)}$.
The direct and decomposed predictions correspond to two different ways of predicting the terminal latent from the same candidate sequence.
The discrepancy between these two terminal estimates can be added as a model-consistency penalty:
\begin{equation}\label{eq:self-consistency}
    C^{(m)}
    =
    C_{\mathrm{goal}}^{(m)}
    +
    \beta
    \left\|
    \hat{\lat}_{t+H}^{(m)}
    -
    \tilde{\lat}_{t+H}^{(m)}
    \right\|_2^2,
\end{equation}
Here, $\beta$ is a hyperparameter that weights the optional self-consistency term.
This penalty does not directly measure whether a candidate action sequence achieves the task.
Instead, it encourages CEM to favor candidates whose terminal predictions are stable under different prefix decompositions.

\section{Experiments}

We evaluate \method{} in terms of planning performance and efficiency, open-loop latent prediction, physical state probing, and ablations.
Experimental settings and implementation details are provided in Appendices~\ref{app:experimental_setup}--\ref{app:runtime_measurement}.

\subsection{Planning Performance and Efficiency}

Table~\ref{tab:main_results} evaluates the planning performance of \method{}.
Under the same planning protocol, \method{} matches or outperforms LeWM in all four environments, increasing the average success rate from 85.8\% to 90.5\%.
Augmenting candidate scoring with the optional self-consistency term in Eq.~\ref{eq:self-consistency} further increases the average success rate to 92.0\%, suggesting that consistency across different prefix-based terminal estimates provides a useful auxiliary signal for distinguishing reliable candidate sequences.

Table~\ref{tab:efficiency} provides a runtime evaluation under the same CEM budget.
We report both the dynamics-evaluation time, which includes action encoding and latent prediction, and the full CEM solve time.
The latter additionally includes goal and observation encoding, candidate scoring, and other data-processing overhead.
Unlike LeWM, which applies its dynamics predictor sequentially over the planning horizon, \method{} evaluates each candidate sequence using a single action-prefix encoding pass followed by parallel latent prediction.
Both variants of \method{} in Table~\ref{tab:efficiency} reduce dynamics-evaluation time and full CEM solve time compared with LeWM.
In particular, \method{} reduces dynamics-evaluation time from 19.7s to 4.1s and CEM solve time from 33.7s to 16.1s, corresponding to speedups of 4.8$\times$ and 2.1$\times$, respectively.
Overall, the proposed algorithm improves planning success while making the repeated evaluation of candidate action sequences substantially more efficient in CEM-based planning. %

\begin{table}[t]
\centering
\caption{
Planning success rate (\%) across environments. 
    With the same planning objective as LeWM,
    \method{} improves the average success rate from 85.8\% to 90.5\%. 
    Adding the optional self-consistency term to candidate scoring further increases the average success rate to 92.0\%.
}
\label{tab:main_results}
\smallskip
\begin{adjustbox}{max width=\linewidth}
\begin{tabular}{lccccc}
\toprule
Method & Two-Room & Reacher & PushT & Cube & Avg. \\
\midrule
PLDM~\citep{sobalstress}
& 97 & 78 & 78 & 65 & 79.5 \\
DINO-WM~\citep{zhoudino}
& \textbf{100} & 79 & 74 & \textbf{86} & 84.8 \\
LeWM~\citep{maes2026leworldmodel}
& 87 & 86 & \underline{96} & 74 & 85.8 \\
\midrule
\method{}
& \underline{98} & \underline{88} & \underline{96}
& 80 & \underline{90.5} \\
\method{} + Self-Consistency
& \underline{98} & \textbf{90} & \textbf{98}
& \underline{82} & \textbf{92.0} \\
\bottomrule
\end{tabular}
\end{adjustbox}

\vspace{-3mm}
\end{table}

\begin{table}[t]
\centering
\caption{
Planning efficiency under the same CEM budget.
Dynamics time includes action encoding and latent prediction, whereas CEM  time covers the complete planning procedure. 
All runtimes are measured on a single NVIDIA 4090.
}
\label{tab:efficiency}
\smallskip
\begin{tabular}{lccc}
\toprule
Method & Model calls & Dynamics time(s) & CEM time(s) \\
\midrule
LeWM        & 5 & 19.7 & 33.7\\
\method{} + Self-Consistency & 3 & 11.9 & 25.7\\
\method{}   & \bf 1 & \textbf{4.1} & \textbf{16.1} \\

\bottomrule
\end{tabular}
\vspace{-3mm}
\end{table}

\subsection{Open-Loop Latent Prediction}
We first quantitatively evaluate open-loop prediction accuracy in latent space. 
Given an initial observation and a sequence of future actions, each model predicts the latent states along the resulting open-loop trajectory.
We compare these predictions with the latent representations encoded from the corresponding ground-truth future observations.

We evaluate predictions over 50 environment steps.
As an action skip of 5 is used, this evaluation corresponds to 10 latent transition steps.
With a maximum prediction horizon of \(H=5\), \method{} predicts each five-step latent segment in parallel and chains two such segments to cover the full evaluation horizon.
In contrast, LeWM applies its one-step transition model autoregressively throughout all 10 latent transition steps.
We use the same set of test trajectories for both methods.

\begin{figure}[!htb]
  \centering
  \includegraphics[width=.9\textwidth,trim=0 0 0 10mm,clip]{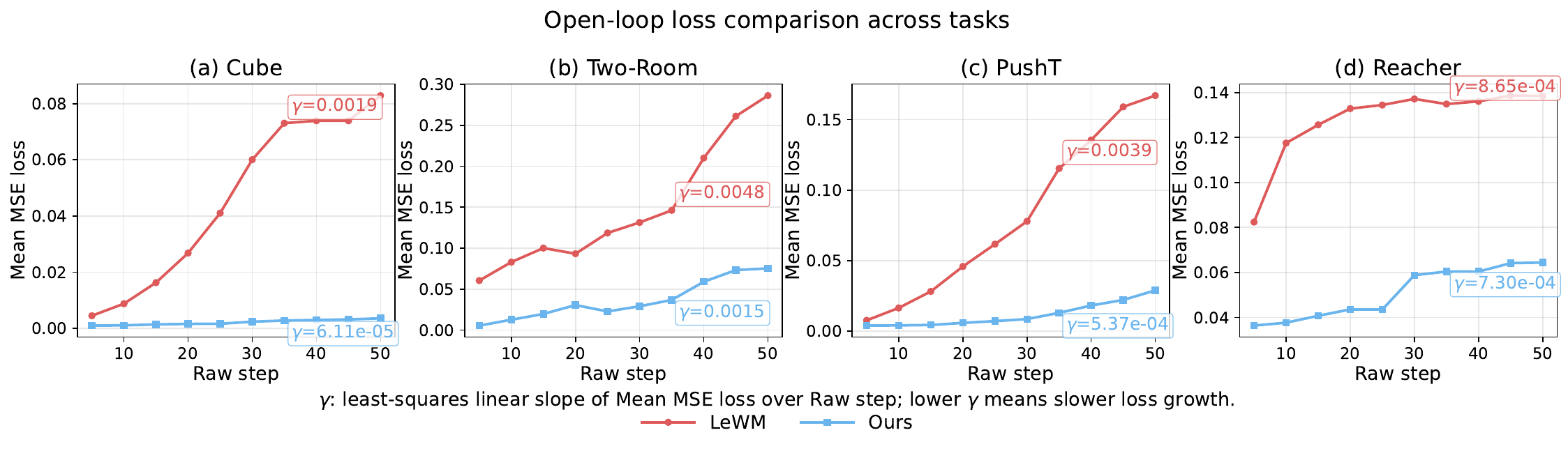}
  \vspace{-3mm}
  \caption{
  Open-loop latent prediction loss over 50 raw environment steps across four tasks.
  The loss is computed between the predicted latent and the latent encoded from the corresponding ground-truth frames.
  The fitted linear slope  $\gamma$ measures the average growth rate of latent prediction error with respect to the prediction horizon.
  Compared with LeWM, \method{} achieves lower loss at every evaluated step and smaller fitted slopes across all four tasks.
  This shows that the action-prefix design substantially mitigates open-loop error accumulation and provides a more error-resistant interface for modeling continuous state transitions over multiple horizons.
  }
  \label{fig:quantity}
  \end{figure}

As shown in Fig.~\ref{fig:quantity}, \method{} achieves substantially lower latent prediction error at every evaluated step across all four tasks.
To quantify how rapidly prediction error increases over the open-loop trajectory, we fit a linear trend
to each loss curve in Fig.~\ref{fig:quantity} using least squares: $\mathcal{L}_{\mathrm{latent}}(t)
\approx
\alpha + \gamma t$, where \(t\) denotes the horizon in raw environment steps, \(\gamma\) measures the average slope of error growth, and $\alpha$ is the intercept.
Across all four tasks, \method{} yields a smaller fitted slope than LeWM.
These results show that action-prefix prediction improves open-loop prediction accuracy and slows the accumulation of latent error over extended horizons.

\begin{figure}[!htb]
  \centering
  \includegraphics[width=.9\textwidth]{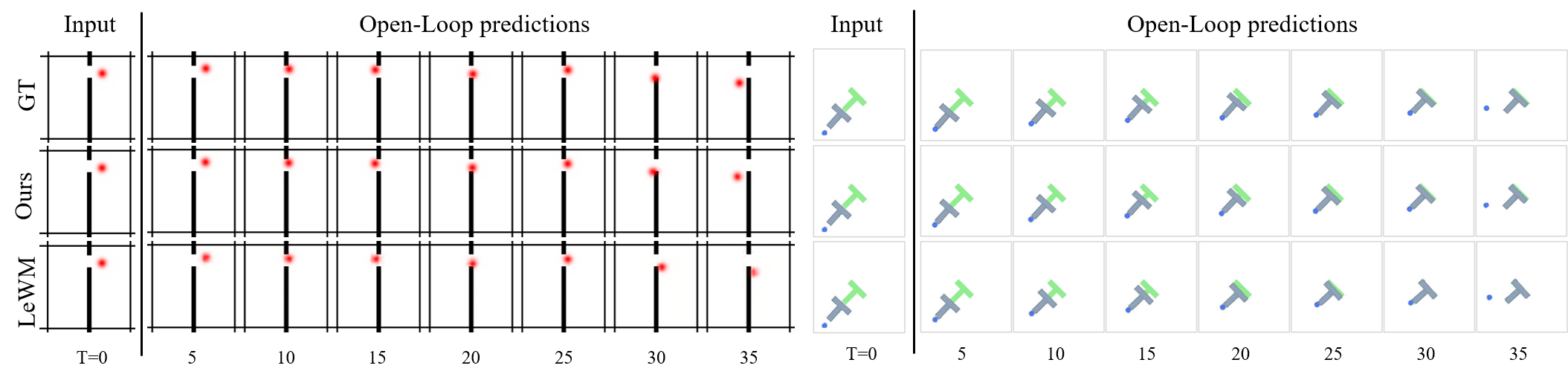}
  \vspace{-3mm}
  \caption{
  Qualitative comparison of open-loop latent predictions, visualized by decoding the predicted latents.
  Both methods are initialized from the same observation and conditioned on the same future action sequence.
  The predictions are similar at short horizons, but the difference becomes more pronounced after 20 environment steps: \method{} remains closer to the ground-truth trajectory, whereas LeWM exhibits increasingly visible drift.
  }
  \label{fig:quality}
  \vspace{-5mm}
\end{figure}

We further provide a qualitative comparison by visualizing predicted latents with a decoder similar to LeWM.
Both models are conditioned on the same initial observation at $t=0$ and the same future action sequence.
As shown in Fig.~\ref{fig:quality}, the decoded predictions of \method{} remain more consistent with the ground-truth trajectory, whereas LeWM exhibits increasingly visible drift over longer horizons due to accumulated rollout error.

\subsection{Physical State Probing}
To examine the physical information encoded in the learned latent representations, we conduct probing experiments on PushT following the protocol of LeWM~\citep{maes2026leworldmodel}. 
We freeze the visual encoder and train lightweight probes to predict three task-relevant physical variables from latent embeddings: agent location, block location, and block angle. 
We report both linear and MLP probes, where the linear probe measures direct accessibility of the physical information, while the MLP probe evaluates whether the information is retained in a nonlinearly recoverable form.

Table~\ref{tab:pusht_probe} reports the mean squared error and Pearson correlation coefficient \(r\).
Under linear probing, \method{} performs similarly to or slightly better than LeWM for agent and block locations, while LeWM more accurately recovers the block angle, indicating that its latent representation retains substantial linearly accessible physical structure.

The advantage of \method{} becomes more pronounced with MLP probes.
It achieves the lowest MSE and highest correlation for all three physical variables, including a substantial reduction in block-angle error relative to LeWM.
These results indicate that \method{} retains physical state information that may not always be linearly organized but is more accurately recoverable through a lightweight nonlinear probe.
We attribute this to the prefix-level training objective: by forcing the model to predict how the state evolves under different action prefixes, rather than only local one-step state evolution, the latent representation must preserve fine-grained physical variables that influence future motion, contact, and object configuration. 
This encourages the encoder and dynamics module to learn more informative state-aware features rather than representations that only support local one-step changes.

\begin{table}[t]
\centering
\vspace{-3mm}
\caption{Physical latent probing on PushT.
We report MSE and Pearson correlation \(r\) for both probes.
}
\label{tab:pusht_probe}
\smallskip
\begin{adjustbox}{max width=\linewidth}
\begin{tabular}{llcccc}
\toprule
\multirow{2}{*}{\textbf{Property}} & 
\multirow{2}{*}{\textbf{Model}} &
\multicolumn{2}{c}{\textbf{Linear}} &
\multicolumn{2}{c}{\textbf{MLP}} \\
\cmidrule(lr){3-4} \cmidrule(lr){5-6}
& & \textbf{MSE $\downarrow$} & \textbf{$r \uparrow$} 
  & \textbf{MSE $\downarrow$} & \textbf{$r \uparrow$} \\
\midrule

\multirow{3}{*}{Agent Loc.}
& PLDM & 0.090 & 0.955 & 0.014 & 0.993 \\
& LeWM & 0.052 & 0.974 & 0.004 & 0.998 \\
& \textbf{Ours} & \textbf{0.048} & \textbf{0.976} & \textbf{0.001} & \textbf{1.000} \\
\midrule

\multirow{3}{*}{Block Loc.}
& PLDM & 0.122 & 0.938 & 0.011 & 0.994 \\
& LeWM & \textbf{0.029} & 0.986 & 0.001 & 0.999 \\
& \textbf{Ours} & \textbf{0.029} & \textbf{0.987} & \textbf{0.000} & \textbf{1.000} \\
\midrule

\multirow{3}{*}{Block Angle}
& PLDM & 0.446 & 0.745 & 0.056 & 0.972 \\
& LeWM & \textbf{0.187} & \textbf{0.902} & 0.021 & 0.990 \\
& \textbf{Ours} & 0.314 & 0.828 & \textbf{0.009} & \textbf{0.995} \\
\bottomrule
\end{tabular}
\end{adjustbox}
\vspace{-1mm}

\end{table}

\begin{table}[!htb]
\vspace{-2mm}
\centering
\caption{
Ablation study. %
All variants use the same training protocol and CEM budget.
}
\label{tab:ablation}
\smallskip
\begin{adjustbox}{max width=\linewidth}
\begin{tabular}{lccccc}
\toprule
Variant & Two-Room & Reacher & PushT & Cube & Avg. \\
\midrule
LeWM & 87 & 86 & 96 & 74 & 85.8 \\
Long-Action LeWM & 76 & 70 & 80 & 58 & 71.0 \\
LeWM w/ Dense Loss & 90 & 78 & 90 & 58 & 79.0 \\
\midrule
\method{} w/o Dense Loss & 96 & 80 & 90 & 72 & 84.5 \\
\method{} & {98} & {88} & {96} & {80} & \textbf{90.5} \\
\method{} w/o State Token & 94 & 82 & 92 & {80} & \underline{87.0} \\
\bottomrule
\end{tabular}
\end{adjustbox}
\vspace{-4mm}
\end{table}

\subsection{Ablation Study}
We analyze the roles of the prediction interface, dense multi-horizon supervision, and state conditioning in Table~\ref{tab:ablation}.
We first examine whether direct long-horizon prediction can be achieved by simply enlarging the temporal span of LeWM.
The resulting baseline, termed {Long-Action LeWM}, expands the original frame-skipped action encoding from 5 to 25 primitive actions and predicts the terminal latent after the longer action block in a single transition.
Its average success rate drops to 71.0\%, indicating that naively extending the temporal span of the original LeWM transition is insufficient for accurate one-pass long-horizon prediction.

We next study the role of dense multi-horizon supervision in \method{}.
Without dense supervision, \method{} achieves an average success rate of 84.5\%, which is comparable to the 85.8\% of the original LeWM while requiring substantially fewer dynamics calls during planning.
Adding dense prefix-level supervision improves the average success rate to 90.5\%.
This result shows that dense supervision is important for learning the more challenging direct multi-horizon prediction problem induced by the action-prefix interface.

To further examine whether the benefit comes simply from adding multi-step supervision, we apply an analogous dense objective to LeWM.
``LeWM w/ Dense Loss'' receives the same variable-horizon inputs as \method{}, performs $H$ autoregressive rollout steps, and averages the latent prediction losses along the rollout.
Its average success rate decreases to 79.0\%, below the original LeWM.
This contrast suggests that dense multi-horizon supervision is not uniformly beneficial across prediction interfaces.
In \method{}, each horizon is predicted directly from the observed anchor latent and its corresponding action-prefix representation, whereas in autoregressive LeWM later-horizon predictions depend recursively on earlier predicted latents.
The effectiveness of dense supervision therefore depends on how multi-horizon dynamics are modeled.

Finally, removing the state token of the action-prefix encoder also degrades performance.
While the state token introduces little overhead, it provides essential context for interpreting the candidate action sequence.
The same action prefix can produce different outcomes depending on the initial agent position, object configuration, scene geometry, and contact conditions.
Conditioning the prefix encoder on the current state therefore helps disambiguate these effects and improves the prediction of the future latent associated with each action prefix.

\section{Limitations and Future Work}

Although \method{} improves long-horizon latent prediction and substantially slows error growth, it still relies on CEM to optimize action sequences at planning time.
As the planning horizon increases, CEM must search over an increasingly high-dimensional action space, making it difficult to identify effective action sequences.
A promising direction is to combine \method{} with hierarchical planning~\citep{zhang2026hierarchicalplanninglatentworld}, where high-level planning operates over temporally extended subgoals rather than individual actions.
We leave this extension to future work.

\section{Conclusion}

We introduced \method{}, a fast latent world model for visual planning. Instead of relying on autoregressive one-step rollout, \method{} predicts future latents directly from the current state and action-prefix representations.
This design reduces repeated dynamics evaluations, mitigates latent error accumulation, and enables dense supervision over continuous state evolution. Experiments across four planning tasks show that \method{} improves planning success while substantially reducing CEM solve time and open-loop prediction error. These results suggest that action-prefix prediction is an effective and efficient dynamics interface for latent world-model planning.

\clearpage
\bibliographystyle{iclr2025_conference}
\bibliography{main}

\clearpage
\beginappendix

\section{Multi-Seed Planning Results}
\label{app:per_seed_results}

In the main paper, we follow the evaluation protocol of the baselines and use the same random seed as LeWM for direct comparison.
To further evaluate the robustness of planning performance,
we additionally evaluate both LeWM and \method{} over seven random seeds, while keeping the planning protocol identical across methods.
As shown in Table~\ref{tab:multi_seed_summary}, \method{} achieves higher mean success rates on all four tasks, improving the overall average from 81.2\% to 85.7\%.
These multi-seed results are consistent with the main-paper comparison obtained under the original LeWM seed, indicating that our performance gains are not specific to a particular random seed.

\begin{table}[h]
\centering
\caption{
Planning success rate (\%) over seven random seeds.
We report mean $\pm$ standard deviation across seven seeds for each task.
The last column reports the average success rate across the four tasks.
}
\label{tab:multi_seed_summary}
\smallskip

\begin{tabular}{lccccc}
\toprule
Method & Two-Room & Reacher & PushT & Cube & Avg. \\
\midrule
LeWM
& 87.0 $\pm$ 4.9
& 83.4 $\pm$ 5.7
& 87.1 $\pm$ 7.2
& 67.4 $\pm$ 8.6
& 81.2 \\

\method{}
& \textbf{95.4 $\pm$ 3.2}
& \textbf{84.0 $\pm$ 5.9}
& \textbf{91.7 $\pm$ 4.1}
& \textbf{71.7 $\pm$ 8.4}
& \textbf{85.7} \\
\bottomrule
\end{tabular}

\end{table}

\section{Experimental Setup and Implementation Details}\label{app:experimental_setup}

\subsection{Datasets and Shared Protocol}

Unless otherwise specified, we follow LeWM for dataset preprocessing, model training, and CEM-based goal-conditioned evaluation, including image resolutions, optimizer settings, and all remaining training and evaluation hyperparameters.
Experiments are conducted on the same offline datasets from Two-Room~\citep{sobalstress}, PushT~\citep{zhoudino}, Reacher~\citep{tassa2018deepmindcontrolsuite}, and OGBench-Cube~\citep{park2025ogbench}.

\subsection{Model Architecture}

We adopt the same ViT visual encoder as LeWM.
The action-prefix encoder is implemented as a causal Transformer over a sequence of action tokens.
The starting token is obtained by mapping the current latent $\lat_t$ through a two-layer MLP with hidden width 768, while the remaining tokens are embeddings of the future actions $\act_t,\ldots,\act_{t+H-1}$.
The token dimension is 192.
The action-prefix Transformer has 3 layers and 6 attention heads, with per-head dimension 32.
We use sinusoidal positional encoding with exponentially spaced frequencies.

For each horizon \(k\), the latent dynamics predictor takes the current latent \(\lat_t\) and the corresponding prefix token \(\mathbf{p}_{t,k}\) to predict the future latent \(\hat{\lat}_{t+k}\) as introduced in Eq.~\ref{eq:predictor}.
It is implemented as a 6-layer residual MLP in which the prefix representation modulates the latent features through AdaLN-Zero.
The predictor uses a latent dimension of 192, a hidden width of 2048, a modulation width of 768, and a dropout rate of 0.1.
The complete model contains 17.9M parameters, comparable to the 18.0M parameters of LeWM.

\subsection{Training Setup}

We train \method{} with a batch size of 128 on all environments except Cube, for which we use a batch size of 32 to obtain more stable and faster convergence. All models are trained for 10 epochs, following the LeWM protocol. During training, the prediction horizon is determined adaptively from the loaded trajectories and clamped to the range $[1,5]$.

\begin{algorithm}[H]
\caption{Dense Action-Prefix Training for \method}
\label{alg:training}
\begin{algorithmic}[1]

\REQUIRE Dataset $\mathcal{D}$, horizon $H$, models
$(\enc,\ape,\pred)$, objective $\trainloss$

\REPEAT

    \STATE Sample
    $(\obs_{t:t+H},\act_{t:t+H-1}) \sim \mathcal{D}$

    \STATE $\lat_{t:t+H}
    \leftarrow
    \enc(\obs_{t:t+H})$

    \STATE $\state_t
    \leftarrow
    \operatorname{MLP}_{\mathrm{state}}(\lat_t)$

    \STATE $\mathbf{x}_{t,0:H}
    \leftarrow
    [\state_t;
    \operatorname{Emb}(\act_{t:t+H-1})]$

    \STATE $[\pre_{t,0},\pre_{t,1:H}]
    \leftarrow
    \ape(
    \mathbf{x}_{t,0:H};
    \mathbf{M}_{\mathrm{causal}}
    )$

    \STATE $\hat{\lat}_{t+1:t+H}
    \leftarrow
    \pred(
    \lat_t,
    \pre_{t,1:H}
    )$

    \STATE Take a gradient step on
    $\nabla_{\theta,\psi,\phi}\trainloss$

\UNTIL{converged}

\end{algorithmic}
\end{algorithm}

\section{Training Procedure}\label{app:training_procedure}

Algorithm~\ref{alg:training} summarizes the training procedure of \method{}. Given a trajectory segment, the shared visual encoder first maps the current and future observations to latent representations. The current latent is transformed into a state token and prepended to the embedded action sequence. A causal Transformer then produces one
representation for every action prefix, and the parallel predictor maps these prefix representations to their corresponding future latents.

Under causal masking, the prefix representation $\pre_{t,k}$ can depend only on the first $k$ actions $\act_t,\ldots,\act_{t+k-1}$. The shared anchor $\lat_t$ and the prefix representations $\pre_{t,1:H}$ are then used to predict all $H$ future latents in one parallel forward pass.

\section{Planning Protocol}\label{app:planning_protocol}

At test time, we follow the same goal-conditioned latent planning protocol as LeWM. We use the same frame-skipped control interface with an action skip of 5 and set the planning horizon to $H=5$. Therefore, each planned action is executed for five primitive environment steps, and the complete planning horizon covers 25 primitive steps. The optimized action sequence is executed according to the same model-predictive control schedule used by LeWM.

For the self-consistency variant, we additionally compare two predictions of the terminal 25-step latent. The first prediction is obtained directly from the complete action prefix. The second is obtained by first predicting an intermediate 10-step latent and then predicting the remaining horizon from this intermediate state. We set the consistency-loss weight to $\beta=1$.

\section{Runtime Measurement}\label{app:runtime_measurement}
 
Runtime is measured on a single NVIDIA RTX 4090 using the same evaluation workload, candidate action sequences, and CEM settings for both methods. Each CEM iteration evaluates 300 candidates and retains 30 elites, with the same number of optimization iterations for both methods. We separately report dynamics-evaluation time and full CEM solve time.

Dynamics-evaluation time covers action encoding and latent prediction, while full CEM time includes all encoding and CEM optimization costs. Both exclude environment stepping and rendering.

Both metrics are accumulated over all planning calls across all evaluation samples and trajectories. The reported runtimes therefore represent the total planning cost of the complete evaluation workload, rather than an average per planning call. This difference in aggregation explains their larger magnitude compared with the averaged planning times reported in LeWM.

\end{document}